Pawan Kumar Singh [1], Shubham Sinha [2],
Sagnik Pal Chowdhury [3], Ram Sarkar [4] and Mita Nasipuri [5]

# Word Segmentation from Unconstrained Handwritten Bangla Document Images using Distance Transform

**Abstract:** Segmentation of handwritten document images into text lines and words is one of the most significant and challenging tasks in the development of a complete Optical Character Recognition (OCR) system. This paper addresses the automatic segmentation of text words directly from unconstrained *Bangla* handwritten document images. The popular Distance transform (DT) algorithm is applied for locating the outer boundary of the word images. This technique is free from generating the over-segmented words. A simple post-processing procedure is applied to isolate the under-segmented word images, if any. The proposed technique is tested on 50 random images taken from *CMATERdb*1.1.1 database. Satisfactory result is achieved with a segmentation accuracy of 91.88% which confirms the robustness of the proposed methodology.

# 1 Introduction

OCR system refers to a process of generating machine editable text when input is given by optical means, like scanning of any text documents either in printed

―――――
**1** Department of Computer Science and Engineering, Jadavpur University, Kolkata, India
pawansingh.ju@gmail.com
**2** Department of Computer Science and Technology, Indian Institute of Engineering Science and Technology, Shibpur, Howrah-711103, India
bitan1994@gmail.com
**3** Department of Computer Science and Technology, Indian Institute of Engineering Science and Technology, Shibpur, Howrah-711103, India
sagnik.pc@gmail.com
**4** Department of Computer Science and Engineering, Jadavpur University, Kolkata, India
raamsarkar@gmail.com
**5** Department of Computer Science and Engineering, Jadavpur University, Kolkata, India
mitanasipuri@gmail.com

or handwritten form. Segmentation is the process of extracting the objects of interest from an image. Segmentation of a document image into its basic entities, *namely*, text lines and words, is considered as a critical dilemma in the field of document image processing. It is a common methodology to segment the text-line which immediately follows the word extraction procedure. The problem becomes more challenging when handwritten documents are considered due to its intrinsic complex nature. Firstly, the writing styles in handwritten documents may be cursive or discrete. In case of discrete handwriting, characters get combined in forming the words. Secondly, unlike machine printed text, handwritten text is not uniformly spaced. Thirdly, ascenders and descenders of the consecutive word images may get frequently connected and these word images can also be present at different orientations. Finally, noise and other artifacts are commonly seen more in handwritten documents than in printed ones. Therefore, if the page segmentation technique is implemented using the traditional two-step approach, the following problems need to be taken care of. In the case of text-line segmentation, major difficulties include the multi-level skewness, presence of overlapping and touching words among the successive text-lines. In the case of word segmentation, difficulties that arise include the appearance of skew and slant at the word-level, the existence of punctuation marks along the text-line and the non-uniform spacing of words which is a common enduring in handwritten documents. So, this two-stage approach to get the words may not be always useful in handwritten recognition because the problems encountered during word segmentation will multiply due to added intricacies in the text-line extraction process. But, if the words are extracted directly from the document page, then these errors can be avoided judiciously. Apart from this, there are situations where text-line segmentation is not essential for example, postal data, medical prescription data, etc.; word segmentation serves as the best alternative for achieving the maximum precision.

From the literature survey performed in [2], it can be seen most of the works addressing on segmentation of handwritten text-line, word or character have been focused on either *Roman*, *Chinese*, *Japanese* or *Arabic* scripts. Whereas a limited amount of work described in the state-of-the-art [3] has been done for unconstrained handwritten *Bangla* documents. S. Saha *et al.* [4] proposed a Hough transform based methodology for text line as well as word segmentation from digitized images. R. Sarkar *et al.* [5] used Spiral Run Length Smearing Algorithm (SRLSA) for word segmentation. The algorithm first segmented the document page into text lines and then smears the neighboring data pixels of the connected components to get the word boundaries which merges the neighboring components and thus word components in a particular text line is extracted. J. Ryu *et al.* [6] considered word segmentation of *English* and *Bangla* texts as a labeling problem. Each gap in a line was labeled whether it was inter-word or intra-word. A normalized super pixel representation method was first presented that extracted a set of candidate gaps in each text line. The assignment problem was considered as a binary quadratic problem as a result of which pairwise relations as well as local properties could be considered. K. Mullick *et al.* [7] proposed a novel approach for text-line segmentation where the image was first blurred out, smudging the gaps between words but preserving space between the lines. Initial segmentation was obtained by shredding the image based on white most pixels in between the blurred out lines. Touching lines were separated by thinning and then by finding most probable point of separation. In [8], initially the contour of the words present in a given text line were detected and then a threshold was chosen based on Median White-Run Length (MWR) and Average White Run-Length (AWR) present in the given text line. After that the word components were extracted from the text lines based on the contour and the previously chosen threshold value. At last, these words were represented in bounded boxes. A few more works [9-13] for word extraction was already done for other *Indic* scripts like *Oriya*, *Devanagari*, *Kannada*, *Tamil*, and *Gurumukhi* handwritten documents. N. Tripathy *et al.* [9] proposed a water-reservoir based scheme for segmentation of handwritten *Oriya* text documents. For text-line segmentation, the document was firstly divided into vertical stripes. Stripe-wise horizontal histograms were then computed by analyzing the heights of the water reservoirs and the relationship of the peak-valley points of the histograms was used for the final segmentation. Based on vertical projection profile and structural features of *Oriya* characters, text lines were segmented into words. For character segmentation, at first, isolated and connected (touching) characters in a word were detected. Using structural, topological and water-reservoir-concept based features, touching characters of the corresponding word images were then segmented. A. S. Ramteke *et al.* in [10] proposed a method based on Vertical Projection Profile to segment the characters from word and extract the base characters by identifying the empty spaces between them. This algorithm was implemented on bank cheques and concludes 97% of efficiency for isolated characters only. H. R. Mamatha *et al.* in [11] described a segmentation scheme for segmenting handwritten *Kannada* scripts into text-lines, words and characters using morphological operations and projection profiles. The method was tested on unconstrained handwritten *Kannada* scripts and an average segmentation rate of 82.35% and 73.08% for words and characters respectively was obtained. S. Pannirselvam *et al.* proposed a segmentation algorithm [12] for segmenting handwritten *Tamil* scripts into text-lines, words and characters using Horizontal and vertical profile. The methodology gave an average segmentation rate of 99%, 98.35% and 96% for text-lines, words and characters respectively. M. Kumar *et al.* described a strip based projection profile technique [13] for segmentation of handwritten *Gurumukhi* script documents into text-lines and a white space and pitch technique for segmentation of text-lines into words. The technique achieved accuracies of 93.7% and 98.4% for text-line and word segmentations respectively. It can be observed from the literature study that for the above mentioned word extraction approaches, the text lines were first considered and then words were extracted from them. But, till date there is no work available for extraction of words directly from unconstrained document images written in *Bangla* script.

## 2  Proposed Work

The proposed work consists of a word segmentation algorithm which directly extracts the text words from the document images without undergoing the error-prone text-line segmentation process. The well-known Gaussian filter [15] is applied to eliminate the noise present in the handwritten document images. The outline of the word images are initially traced by applying DT on the entire document image. This transformation generates an estimation of the word boundaries which is immediately followed by gray-level thresholding. As a result of thresholding, the individual word images are finally extracted from each of the smeared black regions. A suitable measurement is also taken care of to rectify the under-segmentation errors. The details of the present work are discussed in the following subsections.

### 2.1  Distance Transform (DT)

There are a lot of image analysis applications which require the measurement of images, the components of images or the relationship between images. One technique that may be used in a wide variety of applications is the DT or Euclidean distance transform (EDT). The DT maps each image pixel into its smallest distance to regions of interest [14]. Let the pixels within a two-dimensional digital image $I(x, y)$ be divided into two classes – object pixels and background pixels.

$$I(x, y) \in \{0,1\} \qquad (1)$$

The DT of this image, $I_d(x, y)$ then labels each object pixel of this binary image with the distance between that pixel and the nearest background pixel. Mathematically,

$$I_d(x, y) = \begin{cases} 0 & I(x, y) \in \{0\} \\ min(\|x - x_0, y - y_0\|, \forall I(x_0, y_0) \in 0) & I(x, y) \in \{1\} \end{cases} \qquad (2)$$

where, $\|x, y\|$ is some two-dimensional distance metric. Different distance metrics result in different distance transformations. From a measurement perspective, the Euclidean distance is the most useful because it corresponds to the way objects are measured in the real world. The Euclidean distance metric uses the $L_2$ norm and is defined as:

$$\|x, y\|_{L_2} = \sqrt{x^2 + y^2} \qquad (3)$$

This metric is isotropic in that distances measured are independent of image orientation, subject of course to the limitation that the image boundary is digital, and therefore in discrete locations. Fig. 1 shows an example of DT of a binary image. For each pixel in Fig. 1(a), the corresponding pixel in the DT of Fig. 1(b) holds the smallest Euclidean distance between this pixel and all the other black pixels. Illustration of sample handwritten *Bangla* document image and its corresponding EDT are shown in Fig. 2.

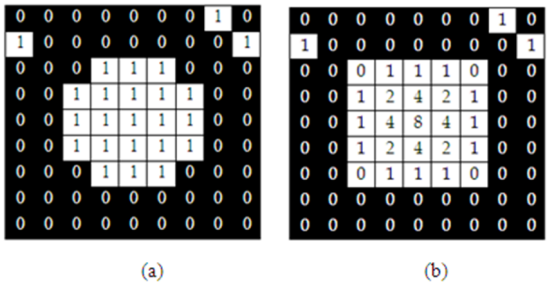

(a)                              (b)

**Figure 1.** Numerical example of EDT. Here, (b) Euclidean distance of each pixel to the nearest black pixel of the input binary image shown in (a). (The distance values are squared so that only integer values are stored)

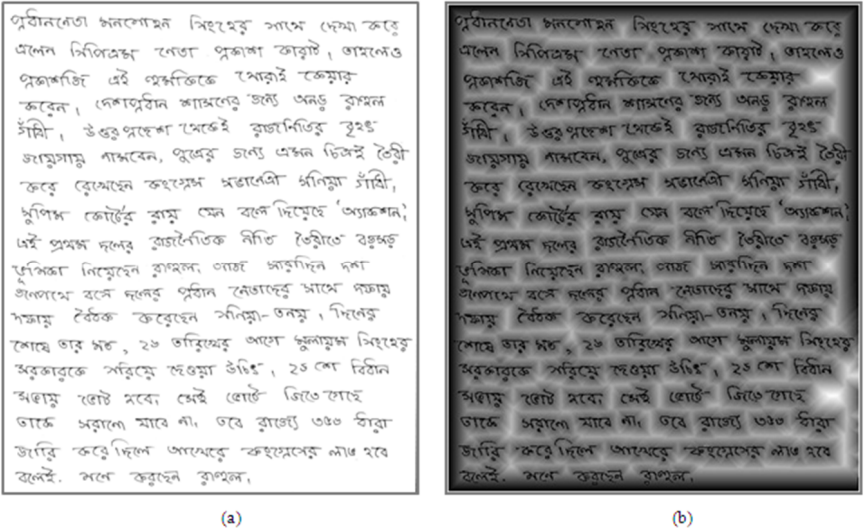

**Figure 2.** Illustration of: (a) sample handwritten *Bangla* document image, and (b) corresponding EDT of (a)

## 2.2 Word Extraction

After performing EDT on the original document image, gray-level thresholding is done to binarize the document images. The value of the threshold $\alpha$ is a user-defined parameter which heavily depends on the handwriting styles and quality of the document image. If the document image incorporates densely packed writing then, the threshold is to be decreased and vice versa. For the present work, the value of $\alpha$ is set after rigorously testing a variety of document images and the optimal value is found to vary between 120 and 180 on a scale of 0-255 gray-level intensities. This thresholding of the document image generates a set of smeared black regions, shown in Fig. 3, which wraps over each of the possible word images. Each smeared black region represents a single group of individual connected components indicating the word image. Now, the word images are formed by applying the connected component labelling (CCL) algorithm [15] in each of these smeared regions.

It is observed that the application of the EDT on the document image yields a non-realistic single connected component along the border region. This single large component includes all the word images which are close to the boundary of the document image. To overcome this, the largest connected component is traversed throughout to locate all the horizontal valleys (present along the vertical outline of the component) and vertical valleys (present along the horizontal outline of the component). These valleys serve as slicing lines to get different words lying along the boundary.

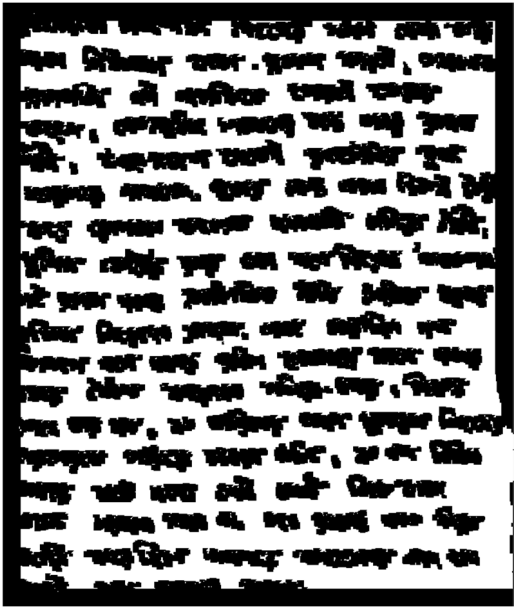

**Figure 3.** Sample handwritten *Bangla* document image showing a smeared black region wrapped over each of the possible word images

## 2.3 Removal of Error cases

In general, there may occur two major errors in any word segmentation method: Over-segmentation and Under-segmentation of the words. If a single word component is erroneously broken down into two/more parts, then it is considered as over-segmentation error. Whereas if two/more words are recognized as a single word, it is considered as under-segmentation error. The present work is free from the over-segmentation errors. This is because of the dynamically selected threshold while binarizing each document image. So, this is one of the key advantages of our technique. But the technique generates some under-segmentation errors. Fig. 4 shows one such situation of under-segmentation for a portion of *Bangla* handwritten document image. Though, under-segmentation errors are found, but the way EDT produces the shape of the smeared regions, it makes the handling of the under-segmentation errors, a bit easier.

The under-segmentation errors are mainly seen of two types: (i) joining of two or more words on the same text-line (i.e., horizontally joined), and (b) joining of two or more words on two successive text-lines (i.e., vertically joined). For the first case, a vertical valley is observed in the word image and this valley is used to separate the joined words. For the second case, the word images protrude to form ascenders and descenders which in turn complicate the situation as the probability of finding an exactly horizontal valley decreases to a large extent. To overcome this, the minimum distance in the horizontal plane is considered and their corresponding boundary points are noted. From the line joining these two points, a user-defined parameter called $\beta$, is taken on both sides, which determines the area of the square through its trajectory. Every two pair of points inside this rectangle is measured and a perpendicular line is drawn from those two points traversing through the boundary of the square. This perpendicular line defines the slicing line between the two vertically under-segmented words. The value of $\beta$ is determined as 0.2 times the word image height.

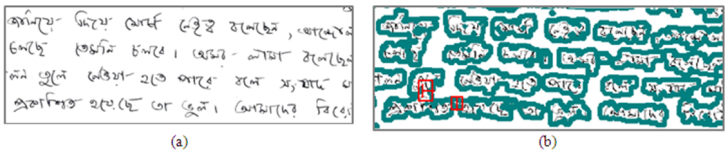

(a)                    (b)

**Figure 4.** Illustration of: (a) a scrap of *Bangla* handwritten document image, and (b) under-segmentation error causing the amalgamation of two text words into a single one

The algorithm of the proposed word segmentation technique is as follows:

| Step 1: | Read the original document image as RGB image and convert it to gray-scale image. |
|---------|-----------------------------------------------------------------------------------|
| Step 2: | Apply Gaussian filter to remove the noise. |
| Step 3: | Perform EDT to the document image. |
| Step 4: | Select a threshold value $\alpha$ for each distance transformed images for performing gray-level thresholding. |

| Step 5: | Locate the boundary points for each binarized word images by examining the 8-connectivity neighborhood of each pixel and then store these boundary points in an array. |
| --- | --- |
| Step 6: | Apply CCL algorithm to label the connected components in the binarized image which, in this case, are the individual word images. |
| Step 7: | Accumulate the pixels that are present in each label into a new array to form an individual word image. |
| Step 8: | Perform a suitable post-processing technique to separate the under-segmented word images. |

# 3  Experimental Results and Analysis

For the experimental evaluation of the present methodology, a set of 50 document pages are randomly selected from *CMATERdb*1.1.1 handwritten database [16]. *CMATERdb*1.1.1 contains 100 document images written entirely in *Bangla* script. For manual evaluation of the accuracy of word extraction technique, we have considered the errors produced due to under- and over-segmented word images. In both the cases, such extracted words are also treated as wrongly extracted words. The performance evaluation of the present technique is shown in Table 1. Fig. 5 illustrates the present word extraction technique on a sample document page.

**Table 1.** Performance evaluation of the present word extraction technique on *CMATERdb*1.1.1 handwritten database

| Database | *CMATERdb*1.1.1 |
| --- | --- |
| Number of document pages | 50 |
| Actual number of words present (T) | 7503 |
| Number of words extracted experimentally | 6894 |
| Number of over-segmented words (O) | 0 |
| Number of under-segmented words (U) | 609 |
| Success rate $[(T - (O + U) * 100)/T]$ | **91.88%** |

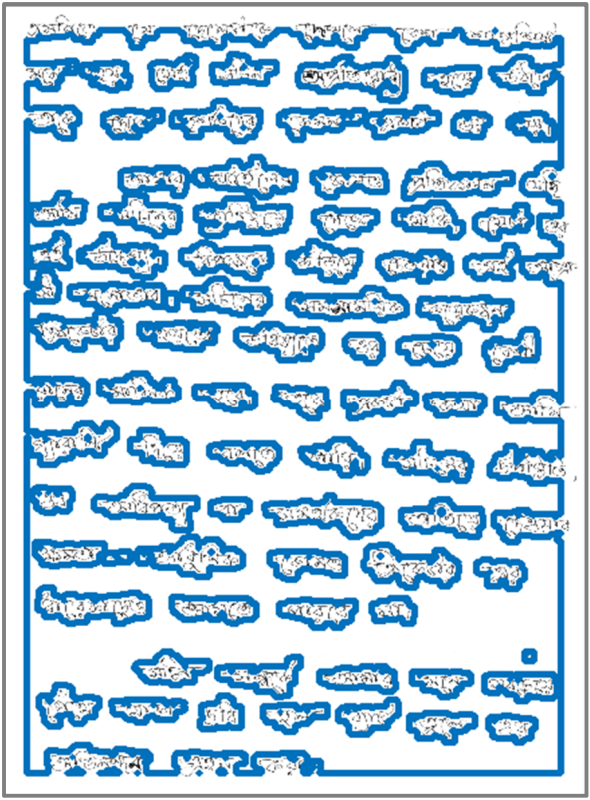

**Figure 5.** Illustration of successful word segmentation technique on a sample document page written in *Bangla* script

# Conclusion

In the present work, we have designed a direct page-to-word segmentation algorithm for unconstrained *Bangla* handwritten document images. A general tendency of the researchers in this domain is to first identify the text-lines and then the extracted text-lines are fed to the word extraction module. Considering the typical complexities of the unconstrained handwritings, we have decided to avoid the text-line extraction module which otherwise could have been generated unwanted errors, cumulative in nature; in due course which would lessen the segmentation accuracy of the word extraction module. Experimental result on the *CMATERdb*1.1.1 handwriting database has shown that the proposed technique yields reasonably satisfactory performance comparable to the state-of-the-art techniques. Along with that, this technique avoids the removal of skew present in the text-lines to some extent as we have extracted the words directly from the document images. Though the results are encouraging, still the proposed technique suffers from under-segmentation issues even if we have tried to rectify these errors to some extent. More appropriate post-processing modules need to be developed to cope up with these situations. Future scope of the work will be to deal with the connected component which is smeared across the border in a better way. Finally, we could say that with minor modifications, this technique could be successfully applied to other *Indic* scripts documents too.

## Acknowledgment

The authors are thankful to the Center for Microprocessor Application for Training Education and Research (*CMATER*) and Project on Storage Retrieval and Understanding of Video for Multimedia (SRUVM) of Computer Science and Engineering Department, Jadavpur University, for providing infrastructure facilities during progress of the work. The current work, reported here, has been partially funded by University with Potential for Excellence (UPE), Phase-II, UGC, Government of India.